\documentclass{ifacconf}
\usepackage{color,amsmath,graphicx,breqn,amssymb,optidef,accents}
\usepackage{algpseudocode,algorithm}
\usepackage{wrapfig, comment}
\usepackage{matlab-prettifier, listings}
\usepackage{booktabs} 
\usepackage{multirow}
\usepackage{natbib}        
\usepackage{bm}
\usepackage{enumerate}
\usepackage{epsfig}
\usepackage{epstopdf}
\usepackage{latexsym}
\usepackage{amsfonts}
\usepackage{url}
\usepackage{xurl}
\usepackage[T1]{fontenc}
\usepackage{verbatim}

\newcommand{\vb}[1]{\lstinline[style=Matlab-editor,basicstyle=\mlttfamily\small,mlscaleinline=false]\!#1\!}

\newcommand{\red}[1]{\textcolor{black}{#1}}
\begin{document}
\begin{frontmatter}

\title{\LARGE \bf Deep Learning of Dynamic Systems using System Identification Toolbox\texttrademark}

\author[First]{Tianyu Dai}
\author[Second]{Khaled Aljanaideh} 
\author[First]{Rong Chen}
\author[First]{Rajiv Singh}
\author[First]{Alec Stothert} 
\author[Third]{Lennart Ljung}

\address[First]{MathWorks, 1 Apple Hill Drive, Natick, MA 01760 USA}
\address[Second]{Aeronautical Engineering Department, Faculty of Engineering, Jordan University of Science and Technology, Irbid, Jordan}
\address[Third]{Division of Automatic Control, Department of Electrical Engineering, Linkoping University, Linkoping, Sweden}

\maketitle
\thispagestyle{empty}
\pagestyle{empty}

\begin{abstract}
MATLAB\textsuperscript{\textregistered} releases over the last 3 years have witnessed a continuing growth in the dynamic modeling capabilities offered by the System Identification Toolbox\texttrademark. The emphasis has been on integrating deep learning architectures and training techniques that facilitate the use of deep neural networks as building blocks of nonlinear models. The toolbox offers neural state-space models which can be extended with auto-encoding features that are particularly suited for reduced-order modeling of large systems. The toolbox contains several other enhancements that deepen its integration with the state-of-art machine learning techniques, leverage auto-differentiation features for state estimation, and enable a direct use of raw numeric matrices and timetables for training models.  
\end{abstract}


\end{frontmatter}

\section{Introduction}
The industry trend of adopting machine- and deep-learning techniques for controls and identification application continues to grow. Latest advancements include the ability to better utilize prior knowledge in configuring the model structure, improve physical interpretability, and use identified models as practical surrogates of large-scale systems. A thorough review of the deep networks is provided in \citep{Pillonetto2023DeepNF}; see also \citep{LJUNG20201175,Chuiso:2019ML}.

The System Identification Toolbox \citep{matlabsitb} aims to accelerate this trend by making it easy to integrate machine-learning inspired non-parametric predictive models, to use neural networks as building blocks of nonlinear dynamic models and to automatically search for and configure the model structure for best generalizability \citep{ALJANAIDEH2021369}. An objective is also to enable the use of identified models for online state estimation and for serving as plant models for nonlinear model predictive control. This paper describes some of the enhancements introduced in the toolbox over the last 3 years (up to the release R2023b \citep{MATLAB23b}).

\begin{itemize}
	\item Section \ref{sec:nlss} describes neural state-space models and their use for reduced-order modeling.
    \item Section \ref{sec:nnbb} describes the construction and use of Nonlinear ARX and Hammerstein-Wiener models using machine-learning regression models and neural networks. 
    \item Section \ref{sec:sindy} describes sparse estimation feature, in particular as they apply to the learning of linear-in-regressor models. 
    \item Section \ref{sec:other} summarizes other significant improvements to the toolbox such as the use of auto-differentiation features in Extended Kalman filters and the ability to work directly with basic MATLAB data types for model training/analysis. 
\end{itemize}

\section{Neural State-Space Models}\label{sec:nlss}
Neural State-Space (NeuralSS) models were introduced in the R2022b release of MATLAB. They provide a way of using neural ordinary differential equations (neural ODE) \citep{RTQChen:2018} for dynamic system representation. In the System Identification Toolbox, they can be viewed as the black-box counterpart of the nonlinear grey-box models that were first introduced in 2007. A NeuralSS model fundamentally represents dynamics in the nonlinear state-space form:
\begin{align} \label{eq:nlss1}
dx(t) &= \mathbf{f}(t,x(t),u(t)) \nonumber \\
y(t) &= \left [ {
	\begin{array}{c}
	y_1(t) \\ y_2(t)
	\end{array} } \right ] = 
	\left [ {
	\begin{array}{c}
	x(t) + e_1(t) \\ \mathbf{g}(t,x(t),u(t)) + e_2(t) 
	\end{array} } \right ]
\end{align}
where $t$ represents the time variable (sample shift in discrete-time case), $x(t) \in \mathbb{R}^{nx}$ are the $nx$ states, $u(t) \in \mathbb{R}^{nu}$ are the model inputs, and $y(t) \in \mathbb{R}^{ny}$ are the measured outputs. $dx(t)$ represents either the derivative $dx/dt$ in the continuous-time case, or the shift operation $x(t+1)$ in the discrete-time case. The states are assumed to be measured variables resulting in the first input group $y_1(t) \in \mathbb{R}^{nx}$, while any additional outputs are grouped under $y_2(t)$. $e_1(t)$ and $e_2(t)$ are the corresponding output noises. No process noise is assumed resulting in an Output-Error structure. As Equation \eqref{eq:nlss1} shows, the model is permitted to be time-varying although the form can be restricted to be time-invariant. The state update function $\mathbf{f}(\cdot)$ and the output function $\mathbf{g}(\cdot)$ are independently parameterized deep feed-forward networks represented by the \vb{dlnetwork} objects in the Deep Learning Toolbox\texttrademark \citep{matlabdl}. These networks are typically composed of a series of fully-connected layers employing \vb{sigmoid}, hyperbolic tangent (\vb{tanh}), or rectified linear unit (\vb{relu}) activations. They are parameterized by the weights and biases constituting the fully connected layers. They can be arbitrarily deep although in practice 2 to 3 hidden layers typically suffice. 
\begin{figure}
\centering
\includegraphics[width=0.9\linewidth]{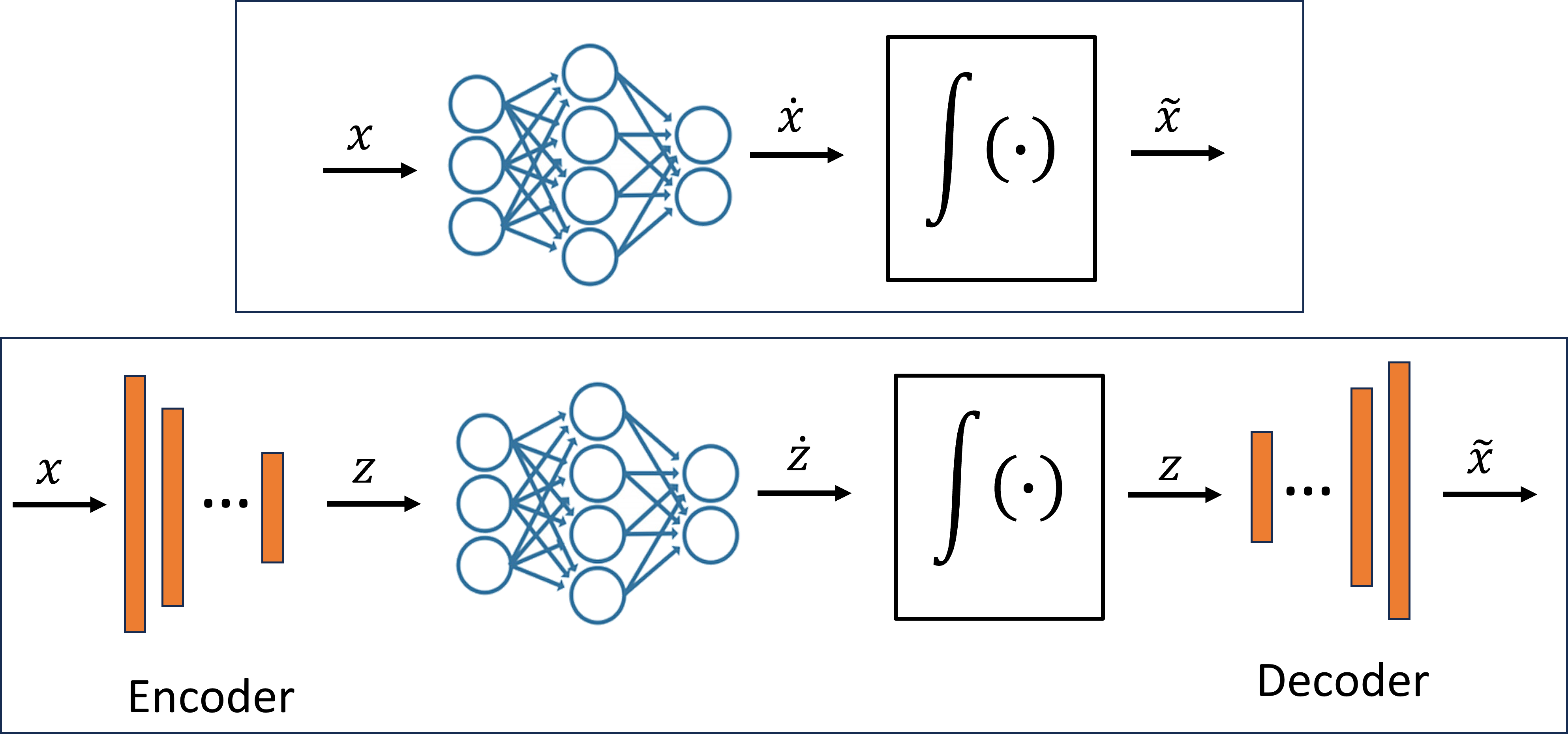}
\caption{Possible configurations of the state transition function of a Neural State Space model. Top: Basic (default) configuration. Bottom: Configuration using auto-encoder to reduce or increase the latent state dimension.}
\label{fig:nlss1}
\end{figure}

NeuralSS models are encapsulated by the \vb{idNeuralStateSpace} objects. One creates a template model by specifying the number of states, and optionally, the number of inputs, number of outputs, sample time and other attributes in the \vb{idNeuralStateSpace} constructor. 
\begin{lstlisting}
% Create a discrete-time neural state-space 
% object with 3 states, 2 inputs, 4 outputs, 
% and sample time of 0.1 seconds.

nss = idNeuralStateSpace(3,NumInputs=2,NumOutputs=4,Ts=0.1)
\end{lstlisting}

The \vb{StateNetwork} and the \vb{OutputNetwork} properties of \vb{nss} contain the deep networks representing the state-transition and the output functions respectively. By default, they employ 2 hidden layers containing 64 \vb{tanh} activations each. The helper function \vb{createMLPNetwork} facilitates easy creation of most commonly used networks. For example,
\begin{lstlisting}
net = createMLPNetwork(nss,"state", ...
  LayerSizes=[4 8 4],Activations="sigmoid");
nss.StateNetwork = net;
\end{lstlisting}
creates a 3-hidden-layer, sigmoid activation based network \vb{net} and assigns it to the \vb{StateNetwork} property of the model, which represents the function $\mathbf{f}(\cdot)$ of Equation \eqref{eq:nlss1}.

The model can be trained using the \vb{nlssest} command whose signature is similar to those used by the other estimation commands in the toolbox such as \vb{ssest} and \vb{nlgreyest}. Suppose \vb{z} is a dataset containing input and output measurements. The weights and biases used by the state-transition and output networks of the model \vb{nss} can be identified by using:
\begin{lstlisting}
nss = nlssest(z, nss, opt)
\end{lstlisting}
Here, \vb{opt} is a training option-set created using the \vb{nssTrainingOptions} command which allows picking the solver, learning rate and other related options. Available solvers are \vb{"ADAM", "SGDM", "LBFGS", "RMSProp"}.

\subsection{Example: Black-Box Model of SI Engine Torque Dynamics}
\begin{figure}[ht]
    \centering
    \includegraphics[width=0.95\linewidth]{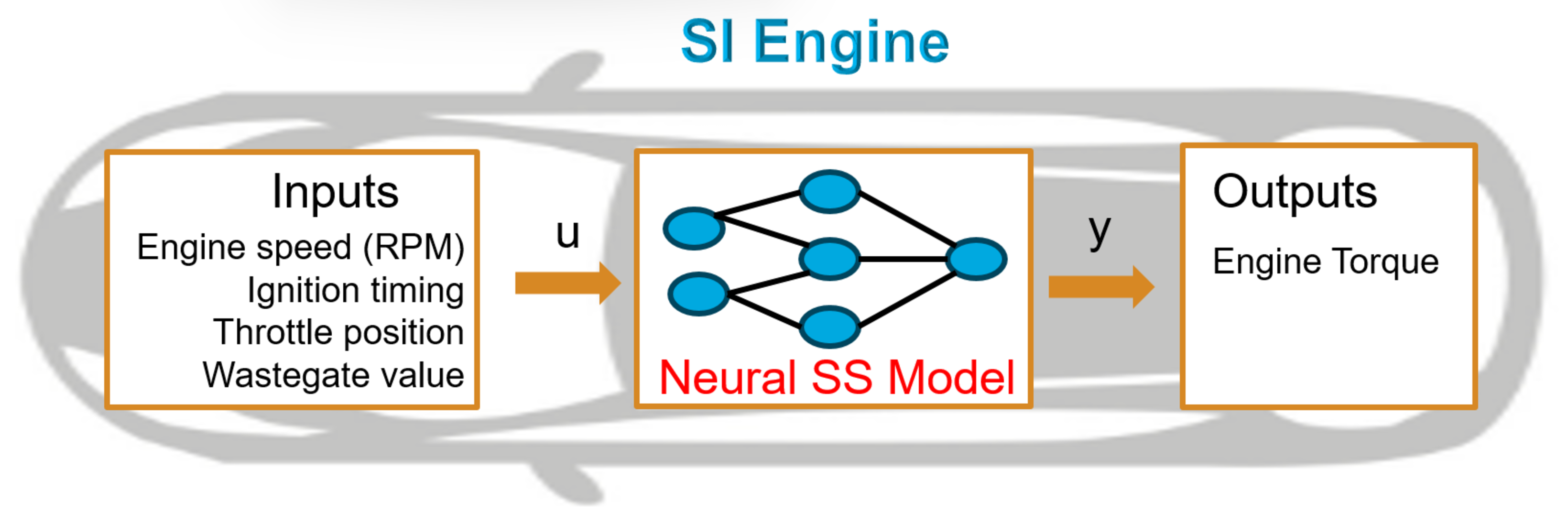}
    \caption{SI Engine model.}
    \label{fig:siengine_model}
\end{figure}

The conventional vehicle reference application, Figure (\ref{fig:siengine_model}), in the MATLAB Powertrain Blockset software represents a full vehicle model with an internal combustion engine, transmission, and associated control algorithms. Embedded in there is a spark ignition (SI) engine model. Our goal is to use simulated signals from this component to create a data-driven proxy of the torque dynamics. 

\begin{figure}[ht]
\centering
\includegraphics[width=0.8\linewidth]{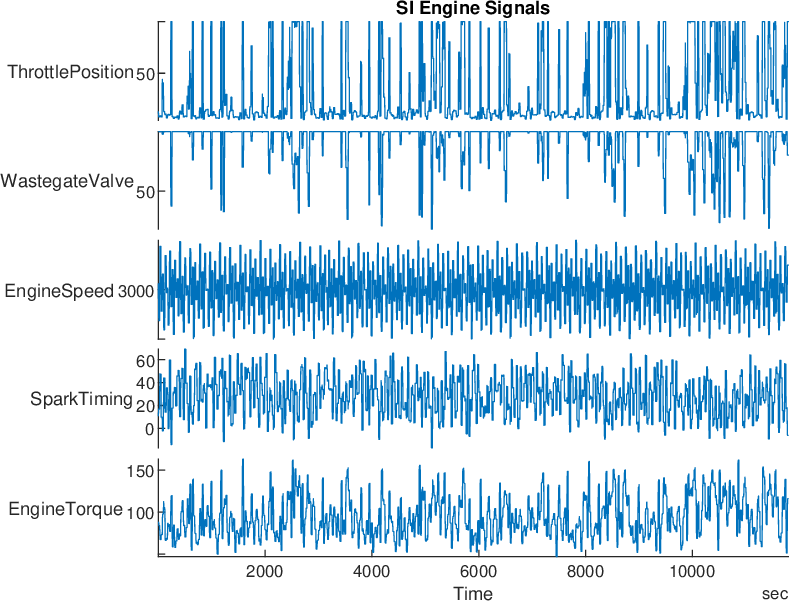}
\caption{SI Engine data.}
\label{fig:SIdata}
\end{figure}

Data for 4 input variables (Throttle position, Wastegate valve, Engine speed, Spark timing) and 1 state/output variable (Engine torque) is collected, as shown in Figure \ref{fig:SIdata}. There are no additional output variables, and hence the output network $g(\cdot)$ is not needed.

To train the model, first create a discrete-time neural state-space object with 1 state, 4 inputs, 1 output and a state network $f(\cdot)$ with 2 layers each of size 128. 

\begin{lstlisting}
nx = 1; % number of states = number of outputs
nssModel = idNeuralStateSpace(nx,NumInputs=4); 

nssModel.StateNetwork = createMLPNetwork(nssModel,"state", LayerSizes=[128 128], WeightsInitializer="glorot", BiasInitializer="zeros", Activations="tanh")
\end{lstlisting}

Then segment the training data (\vb{eData} object that has been downsampled and normalized) into multiple data \red{experiments} to reduce the prediction horizon and improve the training speed. Such segmentations can sometimes lead to more generalizable results. 

\begin{lstlisting}
expSize = 20; 
Expts = segmentData(eData,expSize);
\end{lstlisting}

Next, set up the training options of algorithm. 
\begin{lstlisting}
StateOpt = nssTrainingOptions("adam");
StateOpt.MaxEpochs = 90;
StateOpt.InputInterSample = "pchip"; 
\end{lstlisting}

Finally, train the model using \vb{nlssest} command.
\begin{lstlisting}
nssModel = nlssest(Expts,nssModel,StateOpt)    
\end{lstlisting}

\begin{figure}[ht]
\centering
\includegraphics[width=0.8\linewidth]{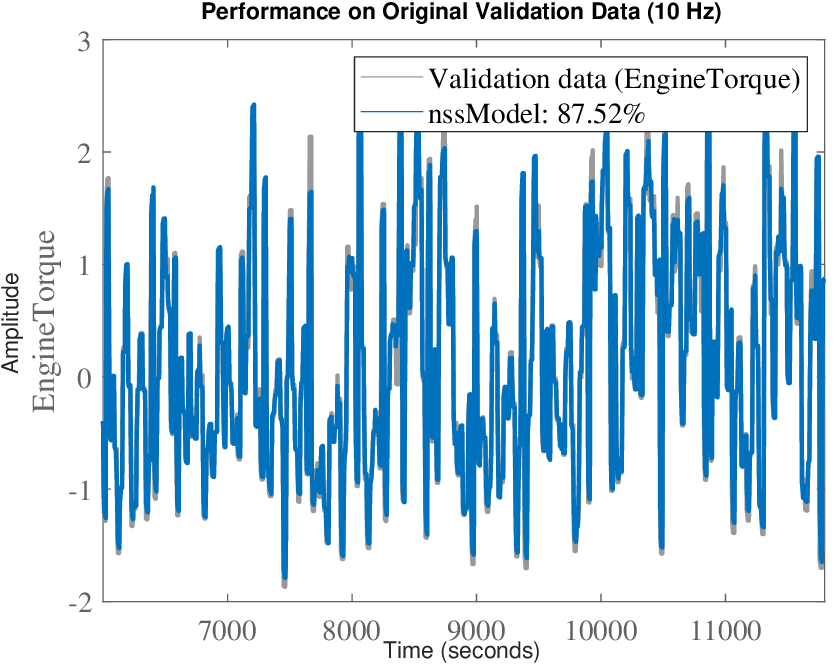}
\caption{SI Engine system: Fit score on validation data using an neural state-space model.}
\label{fig:siengine_score}
\end{figure}

The estimated model's goodness of fit is evaluated using a validation dataset, as depicted in Figure \ref{fig:siengine_score}.
 
\subsection{Example: Feature Reduction Using Auto-Encoders}
This example shows an automatic projection of the model states into a lower-dimensional latent space using auto-encoders. An appealing approach to black-box modeling is to start with a rich set of candidate features and rely on feature extraction techniques such as lasso or principal component regression to obtain a reduced set that actually governs the dynamics.

\begin{figure}[ht]
\centering
\includegraphics[width=0.5\linewidth]{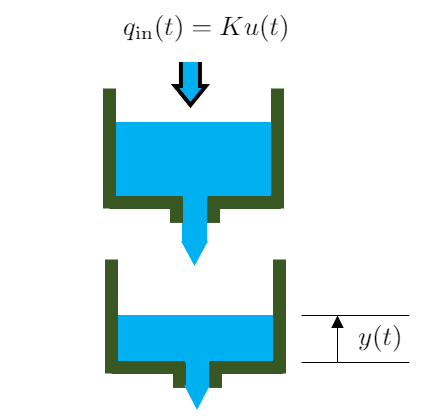}
\caption{Two tank system.}
\label{fig:TwoTankSystem}
\end{figure}

Consider the two-tank system that has been used before to highlight nonlinear identification techniques \citep{ALJANAIDEH2021369}; see Figure \ref{fig:TwoTankSystem}.

Unlike some previous identification approaches for this system \citep{ALJANAIDEH2021369}, we assume that the true model order is unknown and choose a rich set of regressors composed of the lagged I/O variables up to a maximum lag of 10. In a state-space framework, this corresponds to a model of order 20 using the states $x(t) \doteq (y(t-1), y(t-2), \ldots, y(t-10), u(t-1), \ldots, u(t-20))^T$. In the neural state-space structure, we enable the use of an auto-encoder by setting the value of the \vb{LatentDim} property to a finite number; this value indicates the dimension of the latent space.

\begin{lstlisting}
nx = 20; % measured number of states
nu = 1; % number of inputs 
nd = 7; % actual order (latent layer dim)
sys = idNeuralStateSpace(nx, NumInputs=nu, LatentDim=nd); 
net1 = createMLPNetwork(sys, "state", LayerSizes=[], Activations="sigmoid", WeightsInitializer="zeros");

net2 = createMLPNetwork(sys, "encoder",...
   LayerSizes=10, Activations="tanh");

net3 = createMLPNetwork(sys, "decoder",...
   LayerSizes=10, Activations="tanh");
  
sys = setStateNetwork(sys,net1);
sys.Encoder = net2;
sys.Decoder = net3;
\end{lstlisting}

Next, set up the training options. 
\begin{lstlisting}
opt = nssTrainingOptions("adam");
opt.LearnRate = 0.005;
opt.MaxEpochs = 1000;
opt.LossFcn = "MeanSquaredError";
\end{lstlisting}

Finally, use the \vb{nlssest} command to train the model.

\begin{lstlisting}
sys = nlssest(data,sys,opt);
\end{lstlisting}

\begin{figure}
\centering
\includegraphics[width=0.8\linewidth]{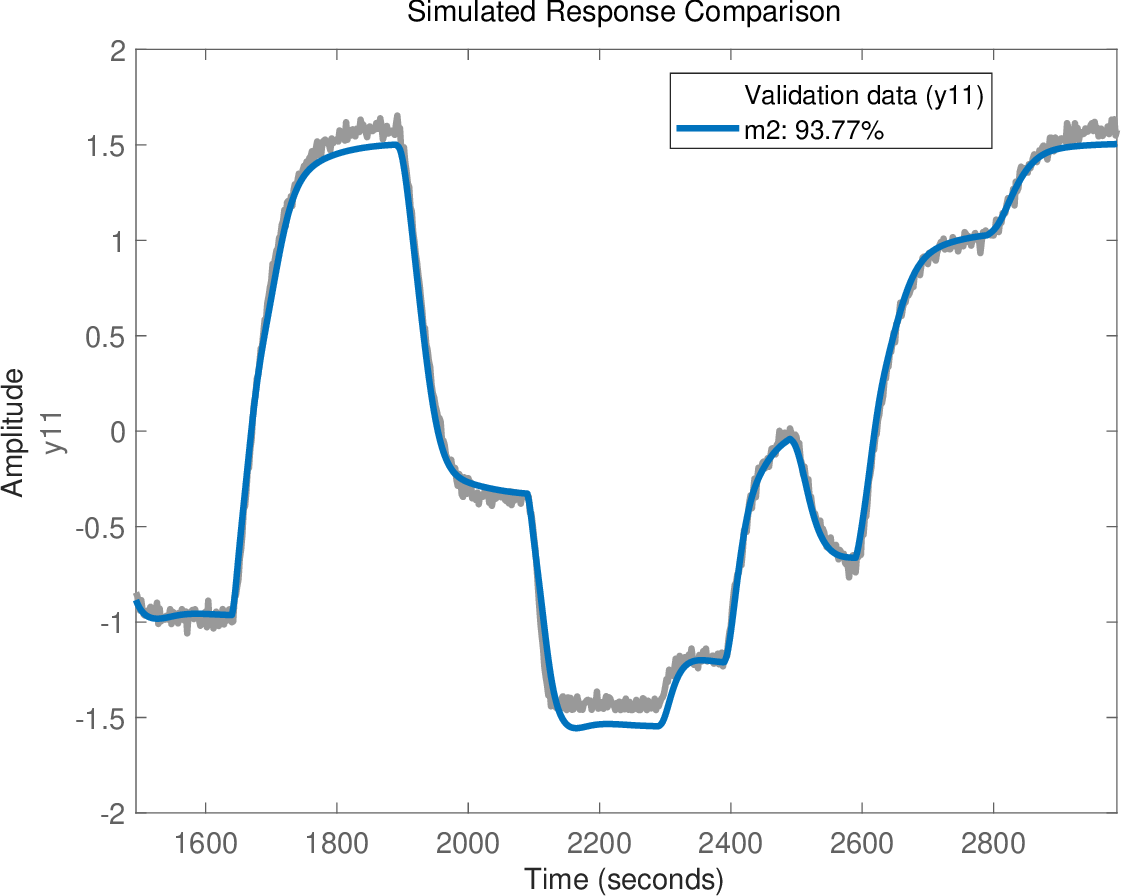}
\caption{Two-tank system: Fit to validation data using an auto-encoder based neural state-space model of latent dimension 7.}
\label{fig:twotank_EN_compare}
\end{figure}

The model \vb{sys} uses 7 states. Figure \ref{fig:twotank_EN_compare} shows the performance of the model on the validation dataset. 

\section{Neural Networks for Nonlinear ARX and Hammerstein-Wiener models}\label{sec:nnbb}
\begin{figure}
\centering
\includegraphics[width=0.9\linewidth]{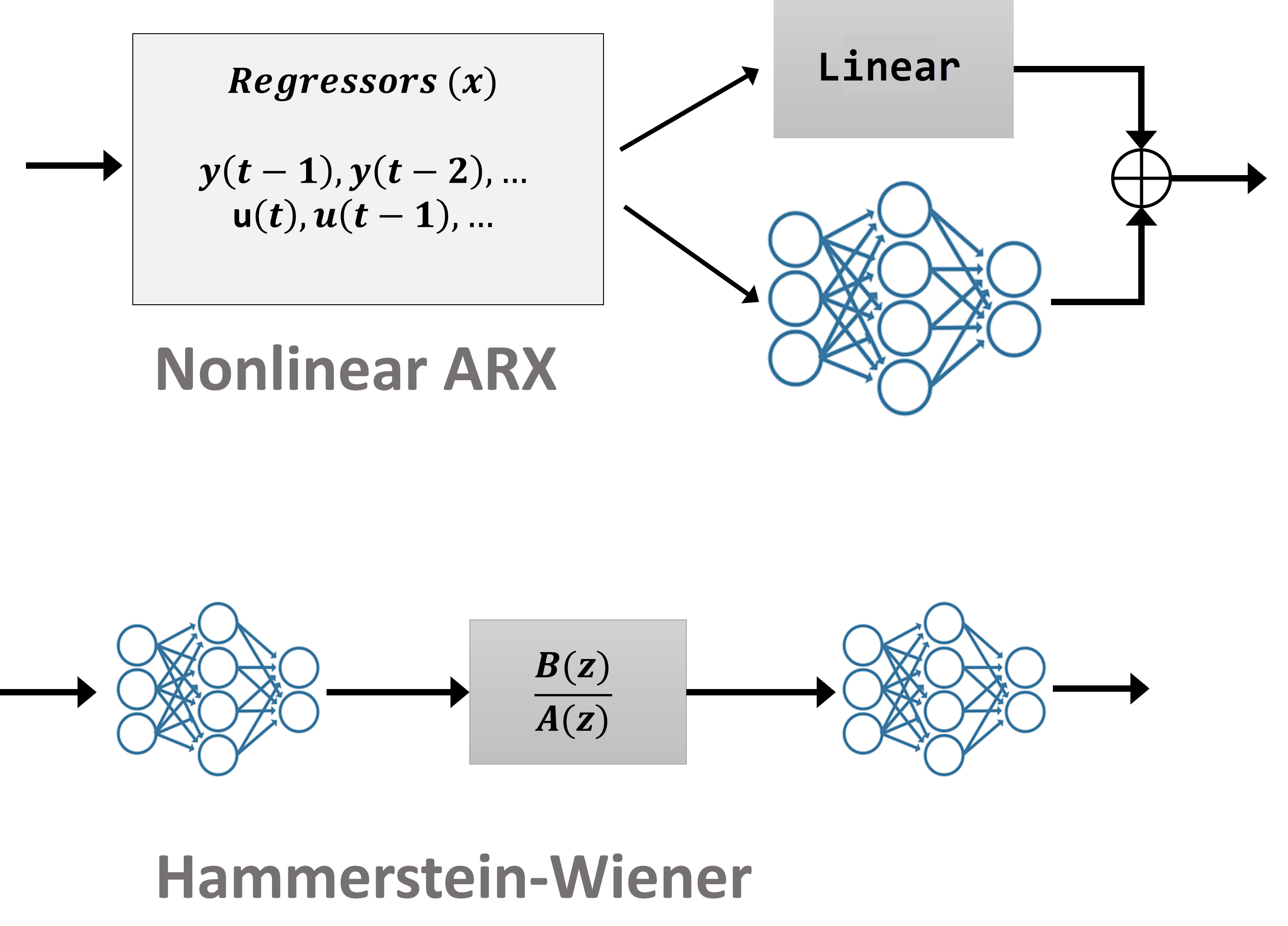}
\caption{Neural network based nonlinear black-box models.}
\label{fig:nlarxhw}
\end{figure}

Nonlinear ARX and Hammerstein-Wiener models are two of the most popular nonlinear black-box structures. They can be viewed as engineering-friendly alternatives to generic recurrent neural networks in that they allow physical reasoning and rather easy deployment for simulations and control design. In the System Identification Toolbox, they are represented by the \vb{idnlarx}, and \vb{idnlhw} objects respectively, and are trained using the commands \vb{nlarx} and \vb{nlhw} respectively. These structures have been upgraded to support the use of Gaussian Process (GP), Support Vector Machine (SVM), and boosting/bagging tree ensembles as their regression functions. The most recent update \red{allows} the use of neural networks to represent the nonlinear input-to-output, or regressor-to-output mappings \red{(Figure \ref{fig:nlarxhw})}. The \vb{idNeuralNetwork} object facilitates the use of neural regression networks (\vb{RegressionNeuralNetwork} object; see\\
\vb{fitrnet}) from the Statistics and Machine Learning Toolbox\texttrademark, and the deep networks (\vb{dlnetwork}) from the Deep Learning Toolbox. 

\subsection{Example: Black-box Modeling of Robot Arm Dynamics}
\begin{figure}
\centering
\includegraphics[width=0.8\linewidth]{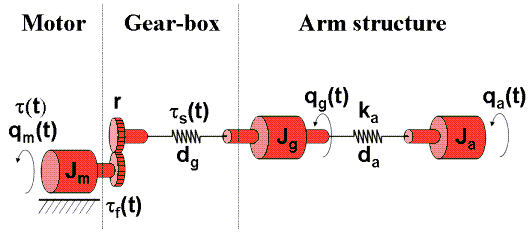}
\caption{Robot arm system.}
\label{fig:robotarm}
\end{figure}

\begin{figure}
\centering
\includegraphics[width=0.9\linewidth]{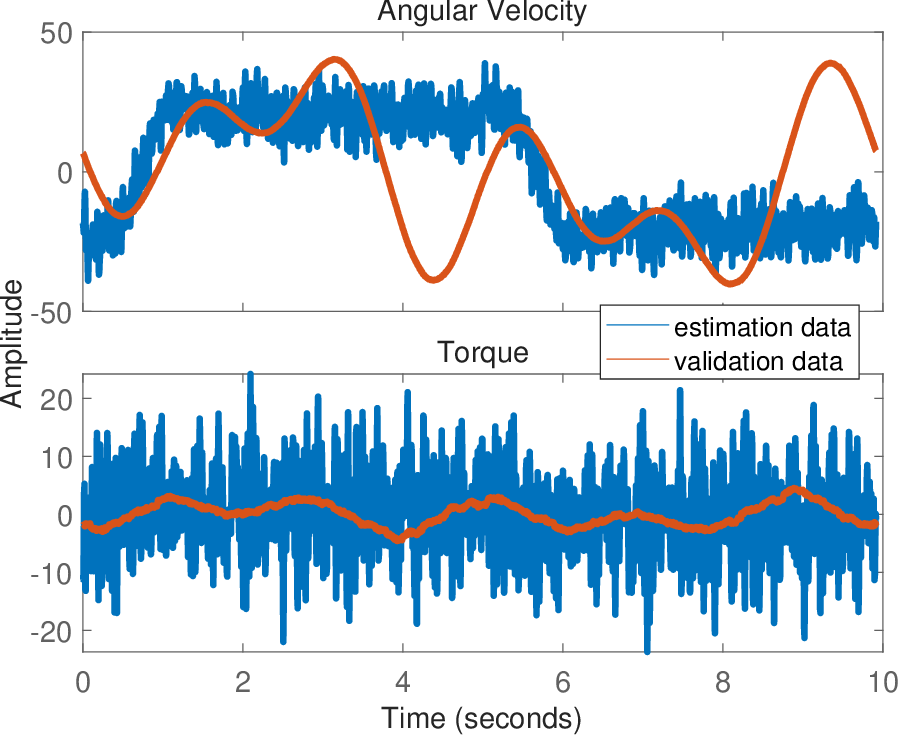}
\caption{Robot arm training and validation datasets.}
\label{fig:robotarm_data}
\end{figure}

Consider a robot arm described by a nonlinear three-mass flexible model (Figure \ref{fig:robotarm}). The input to the robot is the applied torque $u(t) = \tau(t)$ generated by an electrical motor, and the resulting angular velocity of the motor $y(t)=\dot{q}_m(t)$ is the measured output.
This system is excited with inputs of different profiles and the resulting angular velocity of the motor recorded. The training and the validation datasets are shown in Figure \ref{fig:robotarm_data}.

Here \red{are} the main steps involved in the model training\red{:}
\begin{itemize}
\item To begin, prepare model regressors. Here, we pick 2 lags in the output variable and 8 in the input variable.
\begin{lstlisting}
% declare model variables
vars = ["Angular Velocity", "Torque"];
% create linear regressors
R = linearRegressor(vars,{1:2,1:8});
\end{lstlisting}
\item Create a neural network regression function that uses 2 hidden layers, each containing 5 units of \vb{relU} activation functions. 
\begin{lstlisting}
Fcn = idNeuralNetwork([5 5], "relu");
\end{lstlisting}
\item Use the regressor set \vb{R}, and the nonlinear map \vb{Fcn} to instantiate a Nonlinear ARX model structure. 
\begin{lstlisting}
sys1 = idnlarx(vars(1),vars(2),R,Fcn);
\end{lstlisting}
The weights and biases of the fully connected layers of \vb{Fcn} constitute the unknown parameters of the model \vb{sys1}. We want to estimate them so that the model response matches the training data output signal. 
\item Prepare training options. Use Levenberg–Marquardt search method and choose to minimize the simulation errors. This amounts to training the model in a recurrent setup. Also, use "zscore" as the normalization method for the model's regressors and output.
\begin{lstlisting}
opt = nlarxOptions(SearchMethod="lm",...
  Focus="simulation",Display="on");
opt.NormalizationOptions.NormalizationMethod="zscore";
\end{lstlisting}
\item Finally, train the model using the \vb{nlarx} command. For training, split the estimation data into experiments of 500 samples with no overlap. 
\begin{lstlisting}
FS = 500; % data frame size
FR = FS; % frame rate
eDataSplit = segmentData(eData, FS, FR);
sys1 = nlarx(eDataSplit, sys1, opt);
\end{lstlisting}
\end{itemize}

We can similarly train a Hammerstein-Wiener model that employs neural networks as input and/or output nonlinearity. For the robot arm data, we first train a linear model whose order is automatically chosen. This linear model is then used as a starting point for creating a Wiener model structure. The training is performed using the \vb{nlhw} command.
\begin{lstlisting}
% identify a linear model
linsys = ssest(eData, "best", Ts=eData.Ts);
% create a Wiener model structure
uNL = []; % use no input nonlinearity
yNL = idNeuralNetwork([5 5], "tanh");
sys2 = idnlhw(linsys, uNL, yNL);
% prepare estimation options 
opt = nlhwOptions(SearchMethod="lm");
% train the model
sys2 = nlhw(eData, sys2, opt);
compare(vData, sys1, sys2) % validate
\end{lstlisting}

\begin{figure}
\centering
\includegraphics[width=0.8\linewidth]{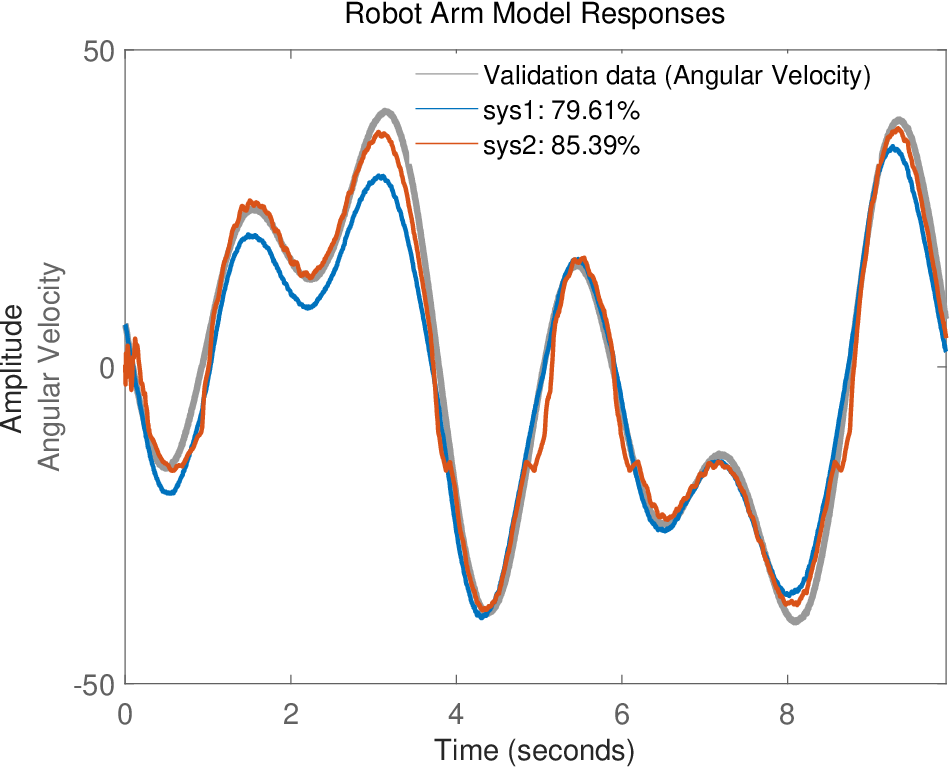}
\caption{Performance of the neural network based nonlinear black-box models on a validation dataset for the robot arm system.}
\label{fig:robotarm_compare}
\end{figure}

\section{Feature Selection and Dictionary-Based Learning} \label{sec:sindy}
For dynamic systems, where measured signals constitute the primary data, Takens' theorem \citep{Takens:1981} motivates the use of lagged variables as regressors. Sets of candidate regressors of various types can be created easily using commands such as \vb{linearRegressor}, \vb{polynomialRegressor}, \vb{periodicRegressor} and \vb{customRegressor}. However, the number of regressors to include is not always obvious. It is then desirable to generate a library (``dictionary") of candidate features and then allow a feature selection algorithm to pick the smallest subset that fits the data. Some popular approaches are Lasso ($\ell_1$ penalty as a relaxation of cardinality requirement), matching pursuit algorithms \citep{matlabwavelet}, and hard thresholding pursuits \citep{yuan2013gradient, fan2009selective}. Some of these can be seen as instances of a proximal gradient descent methods using suitably chosen proximity operators \citep{SINGH:2021_720}.

The R2022b release of MATLAB added \red{the} capability to search for \red{an} optimal subset of regressors of a Nonlinear ARX model based on proximal gradient algorithms. They work by first obtaining an initial model estimate using a full set of regressors and then shrinking the model by sparsifying the set of regressors. In the process, any model structure configurations (such as \vb{RegressorUsage} settings) or estimation options (such as the estimation focus) are honored. The supported sparsity operators are "\red{$\ell_1$}" (for Lasso-like approach), "\red{$\ell_0$}" (for hard-thresholding) and "log-sum" (for the reweighted log heuristics \citep{Candes:2007}).

\subsection{Example: Feature Selection for IC Engine Model}
Consider the problem of modeling the dynamics of an internal combustion engine. The measured dataset contains 1500 samples of the input voltage (volts) and the engine speed output (RPM/100), collected using a sampling \red{interval} of 0.04 seconds. We pick a model order of $na=nb=8$ leading to a model with 16 regressors. Data is split into two portions for estimation (\vb{eData}) and validation (\vb{vData}).

\begin{lstlisting}
% identify a Nonlinear ARX model of order na=nb=8, nk=1
Order = [8 8 1];
sys0 = nlarx(eData,[8 8 1],idSigmoidNetwork);
\end{lstlisting}

Next, perform an estimation that utilizes sparsification options in order to automatically reduce the number of active model regressors.
\begin{lstlisting}
% Specify sparsification options
opt = nlarxOptions(SparsifyRegressors=true, Display="on");
opt.SparsificationOptions.SparsityMeasure = "log-sum";
opt.SparsificationOptions.Lambda = 0.1;
sys1 = nlarx(eData,[8 8 1],idSigmoidNetwork, opt);
\end{lstlisting}

\begin{figure}
\centering
\includegraphics[width=0.9\linewidth]{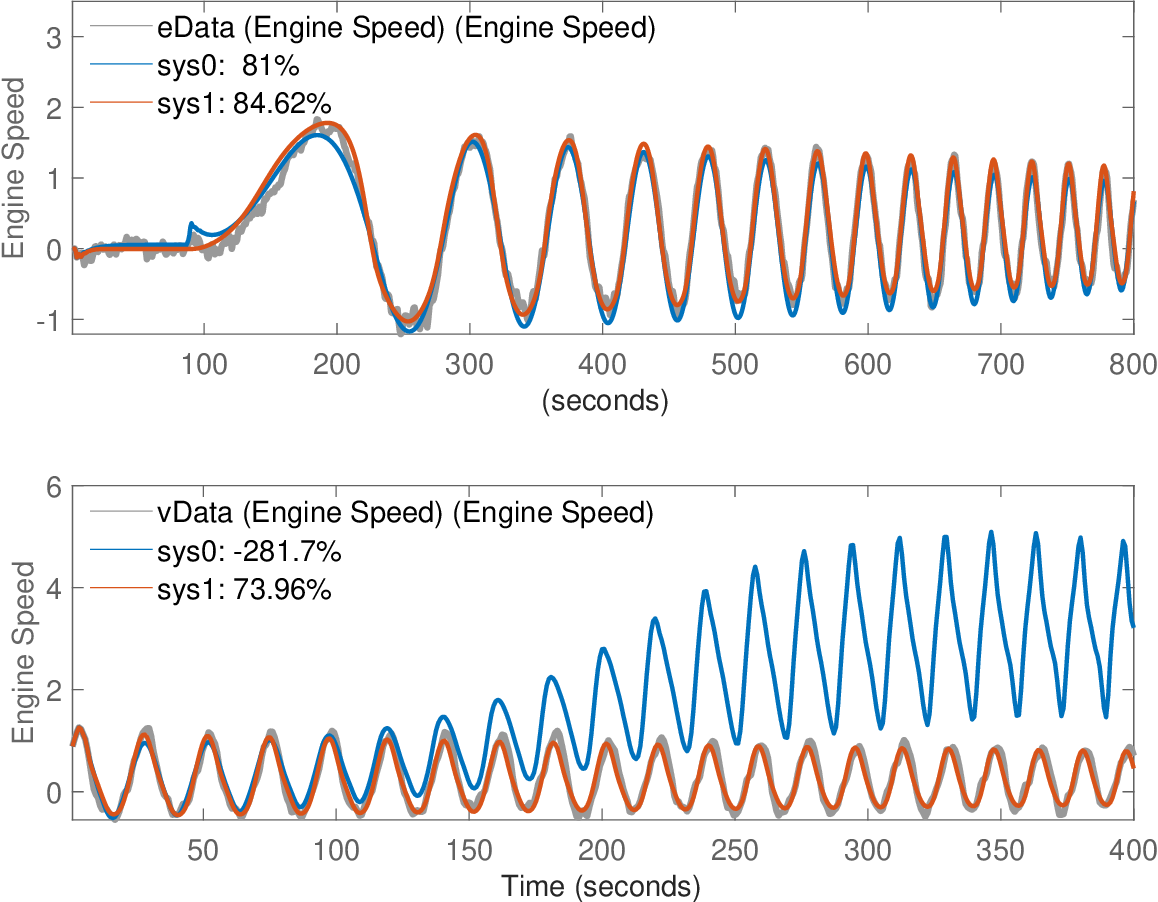}
\caption{Nonlinear ARX model response compared to the estimation data (top) and the validation data (bottom). \texttt{sys0} uses 16 regressors while \texttt{sys1} uses 10 regressors.}
\label{fig:icEngine_compare}
\end{figure}

Compare the simulation performance on the estimation and the validation datasets. 
\begin{lstlisting}
compare(eData, sys0, sys1)
compare(vData, sys0, sys1)
\end{lstlisting}
As Figure \ref{fig:icEngine_compare} shows, the sparsification has notable improvement in the model's quality as attested by the fit to the validation dataset.

\section{Other Improvements} \label{sec:other}
\subsection{Use of Auto-Differentiation in Extended Kalman Filters}
System Identification Toolbox provides tools for recursive/online state estimation. This capability is available in MATLAB (\vb{extendedKalmanFilter}, \vb{unscentedKalmanFilter},
\vb{particleFilter}), as well as in Simulink\textsuperscript{\textregistered} as part of the Estimators library for the toolbox.

\noindent Starting release R2023a, in the Extended Kalman Filter estimator, one can use automatic differentiation (\textit{autodiff}) techniques to generate the Jacobian functions of the state transition and measurement functions. Previously, to specify custom analytical Jacobian functions, one had to write these functions manually.

\subsection{Time-Domain Data Format}
\noindent All the System Identification Toolbox functions that consume data now support the time-domain data to be provided in any of the following formats:
\begin{enumerate}
    \item Double input, output matrix pair. 
    \item \red{Timetable} of scalar valued double variables.
    \item \vb{iddata} object.
\end{enumerate}

\noindent For example, let \vb{u} be a $100\times2$ matrix containing two input signals. Let \vb{y} represent the matrix of corresponding output signals with size $100\times 3$. Let this data be sampled at 10 Hz with a start time of 0 seconds. Consider the following calls for a discrete-time transfer function estimation:
\begin{lstlisting}
np = 2; % number of poles; 
nz = 1; % number of zeros
sys1 = tfest(u,y,np,nz,Ts=0.1);      % double format
VarNames = ["u"+(1:2), "y"+(1:3)]';
TT = array2timetable([u,y],...
  TimeStep=seconds(0.1),StartTime=seconds(0),...
  VariableNames=VarNames);
sys2 = tfest(TT,np,nz,Ts=0.1,...
  InputName=["u1","u2"]);           % timetable format
Z = iddata(y,u,0.1);
sys3 = tfest(Z,np,nz,Ts=0.1);       % iddata format
\end{lstlisting}

\noindent \vb{sys1, sys2, sys3} are identical transfer function models. Clearly, the use of an \vb{iddata} object provides the most systematic way of handling data and sampling information. However, the use of doubles or timetables is \red{sometimes} desirable since they are built-in datatypes in MATLAB and can be used across multiple toolboxes that work with time-domain signals.

\section{Final Comments}
This paper described some of the recent additions to the System Identification Toolbox that are aimed at leveraging deep learning features, and improve the usability of the product. All the examples used in this paper can be accessed from the MATLAB Central File Exchange: \url{https://www.mathworks.com/matlabcentral/fileexchange/165076-examples-of-deep-learning-of-dynamic-systems}

\bibliography{ifacconf} 
\end{document}